\documentclass[11pt]{article}
\usepackage[utf8]{inputenc}
\usepackage[T1]{fontenc}
\usepackage{times}
\usepackage[letterpaper, margin=1in]{geometry}
\usepackage{cite}
\usepackage{amsmath,amssymb,amsfonts}
\usepackage{algorithmic}
\usepackage{graphicx}
\usepackage{textcomp}
\usepackage{gensymb}
\usepackage{multirow}
\usepackage{adjustbox}
\usepackage{url}
\usepackage{array}
\usepackage{xcolor}
\usepackage{float}
\usepackage{caption}
\usepackage{subcaption}
\usepackage[hidelinks]{hyperref}
\usepackage{fancyhdr}
\fancypagestyle{plain}{%
  \fancyhf{}\fancyfoot[C]{\thepage}}
\newcommand{\IEEEmembership}[1]{}

\providecommand{\appendices}{\appendix}

\renewenvironment{abstract}
  {\vspace{1em}\begin{center}\bfseries Abstract\end{center}%
   \begin{quotation}\noindent\small}
  {\end{quotation}}
\newenvironment{keywords}
  {\vspace{0.5em}\noindent\small\textbf{Keywords:\ }}
  {\vspace{1em}}
\title{\bfseries State of Health Estimation using Convolutional and
Bidirectional LSTM Neural Networks tuned by Bayesian Optimization}
\author{
  Panagiotis Eleftheriadis \and
  Foivos Georgios Kyrgios \and
  Sonia Leva
}
\date{%
  \normalsize Department of Energy, Politecnico di Milano,\\
  via Lambruschini 4a, 20156 Milano, Italy\\[0.4em]
  \small\texttt{panagiotis.eleftheriadis@polimi.it}
}
\begin{document}
\maketitle
\thispagestyle{plain}
\begin{abstract}
In this research, a novel framework is proposed for the SOH estimation, which employs a hybrid deep learning architecture of a concatenation of a Convolution Neural Network (CNN) and a Bidirectional Long Short-Term Memory (BiLSTM) Neural Network (NN) with the integration of Bayesian Optimization-based hyperparameter tuning for the network. Three different deep learning architectures are being evaluated: standalone recurrent models, CNN-RNN architectures and CNN-RNN combinations enhanced with intermediate Fully Connected (FC) layers. Among the three, the model with the intermediate FC layers demonstrated the highest predictive accuracy. A comprehensive feature engineering approach combines capacity (Q), voltage (V), Incremental Capacity Analysis (ICA), and Differential Voltage Analysis (DVA), with systematic evaluation of multiple combinations to identify the optimal input representation.
To validate the proposed method, three publicly available datasets were utilized, ensuring reproducibility of the results, two from external sources and one developed by the author of this study using a unique experimental setup. The comparison study was performed using the Mean Absolute Error (MAE), the Root Mean Squared Error (RMSE) and the FLoating-point OPerations (FLOPs) as evaluation metrics.
\end{abstract}
\section{Introduction}
\label{Introduction}
Given the recent policy directions \cite{fitto55}, the widespread electrification across all sectors, particularly in transportation with the increasing penetration of Electric Vehicles (EVs) into the market, alongside the increasing renewable energy integration, has intesified demand for Energy Storage Systems (ESS). Driven by current trends and advancements in battery materials, Battery Energy Storage Systems (BESS) are gaining increasing prominence across various applications \cite{forecast5030032}. Effective battery management is fundumental to ensuring safe, dependable, and optimal performance of BESS. A key element of any Battery Management System (BMS) is the accurate estimation of the State of Health (SOH), which delivers essential insights into the battery capacity \cite{overviewSOH}. Reliable SOH estimation enables effective scheduling, planning, and overall system optimization. Multiple approaches have been proposed for the SOH estimation and can be categorized as either model-based or data-driven methods. Given the complex internal dynamics of batteries and the growing capabilities of artificial intelligence, deep learning methods, within the data-driven category, have garnered significant attention and exhibited remarkable performance in recent studies \cite{overviewSOH}.
Among data-driven SOH estimation techniques, the methods that have garnered greater interest include RNN as a model type, as well as hybrid approaches. Hybrid approaches encompass the fusion of different deep learning architectures or the incorporation of optimization algorithms alongside data-driven methods, as detailed in \cite{overviewSOH}. RNNs have the ability to leverage information from prior predictions as well as measurements to enhance estimation accuracy.
Due to the remarkable capabilities of current neural networks, deep-learning methods have obtained the ability to efficiently process time-series data, showcasing remarkable performance in estimation and prediction applications. Some of the more established deep-learning SOH estimation methods are RNNs, LSTMs and CNNs, among others. It is worth noting that, LSTMs' ability to capture temporal dependencies in extensive time-series datasets makes it one of the most promising alternatives.
In particular, in \cite{MENG2023109288}, the authors proposed an SOH estimation method utilizing an LSTM network and employed Bayesian Optimization (BO) for tuning the hyperparameters of the model. Through this approach, they effectively captured the patterns of the Li-ion battery ageing procedure and significantly improved the accuracy of predictions. Additionally, Bharath et al. \cite{bharath2024long} introduced a cascaded LSTM-RNN model which demonstrated improved efficiency compared to existing state-of-the-art benchmarks. In \cite{LI2025236342}, a BiLSTM model was proposed and coupled with Particle Swarm Optimization (PSO) for hyperparameter tuning. By implementing this approach, MAE and RMSE were limited to under 2.6\%. Lastly, researchers in \cite{xiaoCNNLSTM} developed a CNN-LSTM hybrid framework, without altering the model architecture and parameters, aiming to create a simplified SOH estimation approach. They managed to maintain the model's robustness while ensuring its adaptability to different battery conditions. The aforementioned studies clearly demonstrate the effectiveness and popularity of deep-learning approaches for SOH estimation. However, despite these advancements, challenges related to computational complexity, real-world implementation, and data scarcity still remain. To overcome these gaps, this paper proposes an automated hyperparameter optimization approach specifically designed for battery SOH estimation.
This research presents an integrated framework based on the combination of a CNN and a BiLSTM incorporating hyperparameter tuning through BO coupled with Gaussian Process (BOGP). The overarching objective is to create a comprehensive model that can reliably forecast the SOH under partial discharge data. A novel set of hyperparameters was introduced while maintaining a compact network architecture, expanding the hyperparameter space to include BiLSTM layer count, units per BiLSTM layer, Fully Connected (FC) layers count, neurons per FC layer, and their activation functions. CNN filter counts across both convolutional layers were also searchable. The Adam optimizer learning rate was included as an additional tunable parameter. This comprehensive hyperparameter search space allows discovery of near-optimal configurations without requiring manual intervention. Nevertheless, network performance depends heavily on the hyperparameter range and selection, representing a potential limitation of this research.
To assess the effectiveness of the approach, the proposed framework was applied to three publicly available datasets were utilized, ensuring reproducibility of the results, two from external sources and one developed by the author of this study using a unique experimental setup, while conducting a performance comparison with existing methods. For each dataset, an exhaustive analysis of input features was conducted to identify the combination that produced the most effective predictive outcomes for each solution.
The primary contributions of this work are:
\begin{itemize}
\item Proposing a structure combining a CNN for the extraction of spatial features, while using in cascade connection a RNN capturing the temporal variations of the battery dynamics.
\item Development of an innovative hyperparameter space allowing for efficient exploration utilizing the Bayesian algorithm to attain optimal performance.
\item Three battery datasets analyzed and used to validate the proposed model.
\item An exhaustive input feature analysis between five combinations identifying the most promising combination.
\end{itemize}
This work is structured as follows: Section II describes the datasets employed and feature engineering approach, Section III details the proposed SOH estimation methodology and model training, Section IV reports the experimental results, Section V provides discussion of findings, and Section VI concludes the research.
\begin{table*}[!htbp]
    \centering
    \caption{Datasets Main Features}
    \label{SOHDatasets}
    \begin{adjustbox}{max width=\textwidth}
    \begin{tabular}{c|c|c|c}
        \hline
        \textbf{Features} & \textbf{Oxford} & \textbf{NASA} & \textbf{PoliMi} \\
        \hline
        Battery Type & 8 740mAh 18650 - NMC {lpb}  & 4 2Ah Li-Ion & 6 LG 2.5Ah 18650 NMC \\
        Testing Equipment & Bio-Logic MPG-205 & - & Neware battery tester \\
        Temperature Range & 40\degree C &  23\degree C & 35\degree C \\
        Stressing Cycles & Artemis & Constant & UDDS, US06, LA92, Mixed1-Mixed6 \\
        Sampling rate & 1 second & 5 seconds & 1 second \\
        \hline
    \end{tabular}
    \end{adjustbox}
\end{table*}
\section{Datasets}
As previously mentioned, three datasets were utilized in this study. The first is the PoliMi dataset, a private dataset obtained through a collaboration between TUB and Politecnico di Milano. The other two datasets, Oxford and NASA, are publicly available. The key characteristics of these datasets are summarized in \autoref{SOHDatasets}. Detailed information about the two public datasets can be found in \cite{DOSREIS2021100081}, while details about the private dataset are provided in \cite{Polimi}.
For all datasets, the features are information related to the Voltage, the Capacity, the ICA curve, and the DVA curve, given by \autoref{eqn:ic} and \autoref{eqn:dv}, respectively. The Incremental Capacity (IC) is defined as:
\begin{equation}\label{eqn:ic}
\mathrm{IC}=\frac{\mathrm dQ}{\mathrm dV}
\end{equation}
where the $dQ$ is the change in capacity and $dV$ is the corresponding change in voltage. Similarly, the Differential Voltage (DV) is calculated as:
\begin{equation}\label{eqn:dv}
\mathrm{DV}=\frac{\mathrm dV}{\mathrm dQ}
\end{equation}
These metrics are plotted against voltage in order to produce the ICA and DVA curves, subsequently processed using Savitzky-Golay filtering, eliminating noise and preparing the data for feature extraction.
\begin{figure}[!h]
    \centering
    \includegraphics[width=0.65\columnwidth]{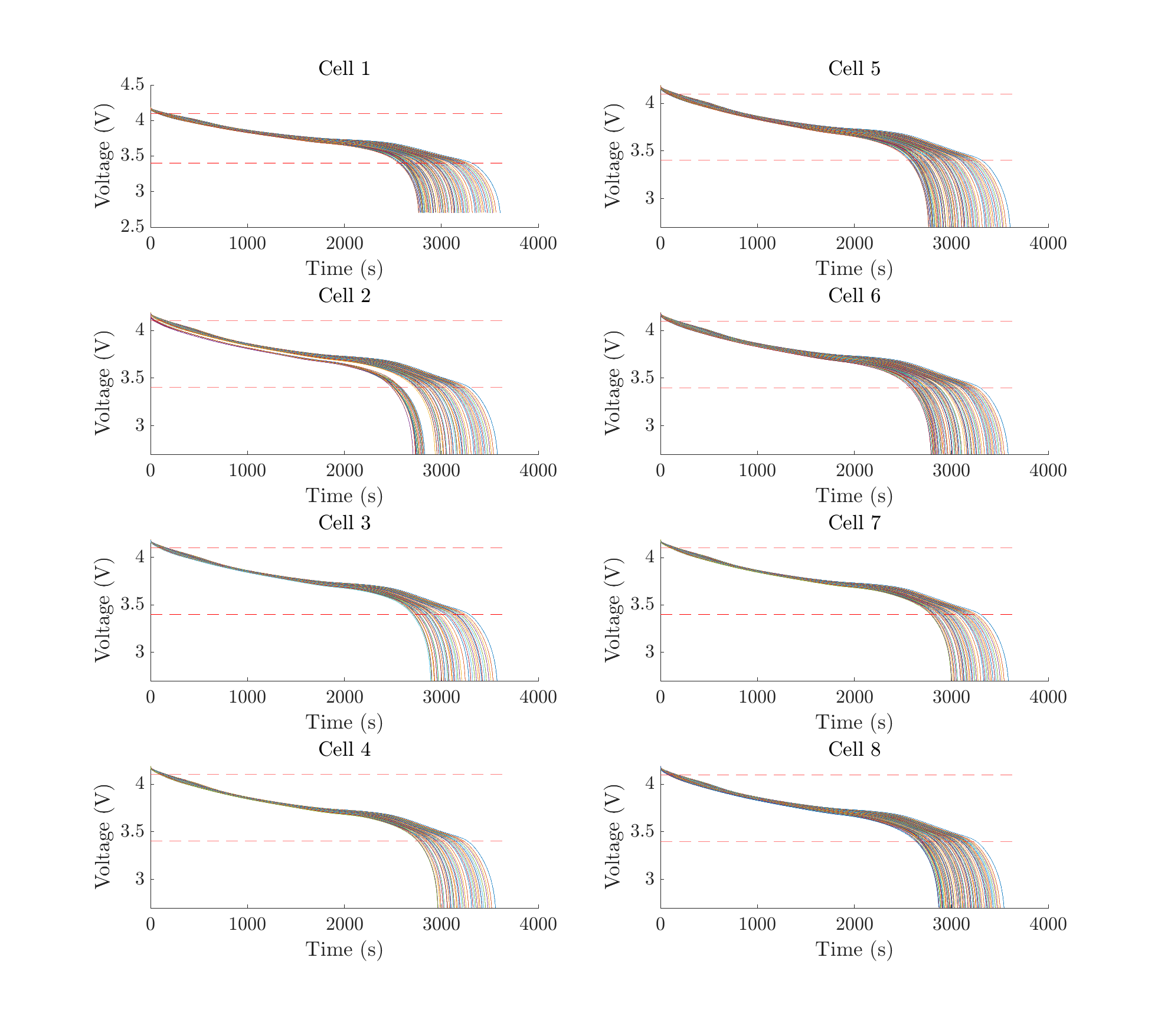}
    \caption{Oxford dataset cell voltage profiles during cycling. Red lines demarcate the voltage window employed for subsequent analysis.}
    \label{ds_fig:1}
\end{figure}
\subsection{Oxford dataset}
\begin{figure}[h]
    \centering
    \includegraphics[width=0.5\columnwidth]{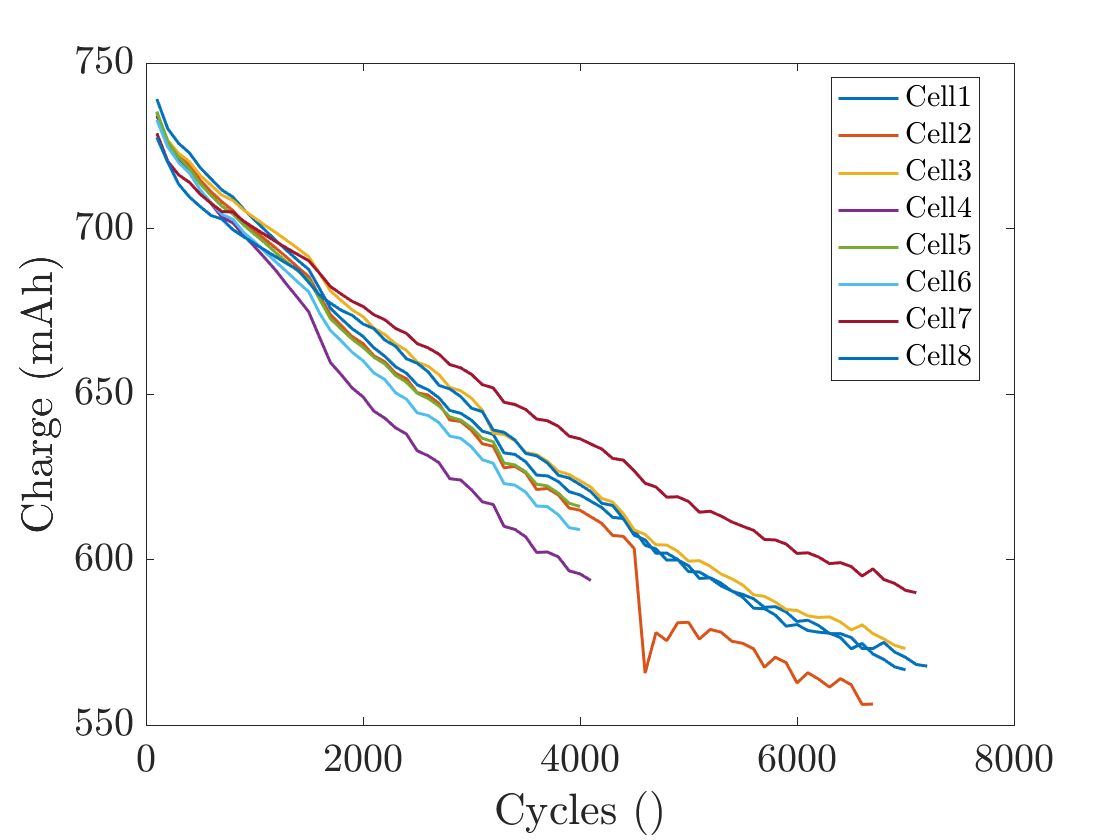}
    \caption{Oxford dataset discharge capacity per cycle. Cells 4 and 6 were disconnected during testing.}
    \label{ds_fig:2}
\end{figure}

For this study, the Oxford University dataset has been utilized. Particularly, eight NMC lithium-ion batteries, connected in parallel, underwent cycling ageing. The testing process was conducted at a constant temperature of 40 \degree C. The upper and lower voltage limits of the batteries were 4.2 V and 2.7 V respectively, while the nominal value was 3.7 V. Additionally, the nominal capacity was 740 mAh. The cycling protocol initiated with a constant-current, constant-voltage charging profile. Subsequently, the batteries were discharged based on the urban Artemis profile. Further information on characterization measurements, charging and discharging rates, and other specifications are provided in the dataset's supplementary file. \autoref{ds_fig:1} depicts all the voltage profiles of the dataset's cells, providing a clear view of the ageing phenomenon through the reduction of discharging time as cycling progresses.
The ageing mechanism can also be observed in \autoref{ds_fig:2} where the maximum charge of every cell gradually drops as the number of cycles increases. For the purposes of this study, only the discharge phase data have been utilized, as these are more practically relevant. Nevertheless, the implementation of this framework could also be based on the charging stage, yielding similar results. In particular, features are extracted from discharge cycles conducted at 1C for the Oxford University dataset, with nominal capacity values referenced against pseudo-OCV measurements. To minimize the processing time required for each battery, instead of considering the full voltage range of 2.7-4.2 V, this method narrows the window of interest to 3.4-4.1 V. This range incorporates the highest peak of the ICA plot and captures most of the useful information \cite{lin2022constant}. This approach provides an optimal framework, minimizing computational costs without compromising data quality.
\subsection{NASA dataset}
\begin{figure}[!h]
    \centering
    \includegraphics[width=0.65\columnwidth]{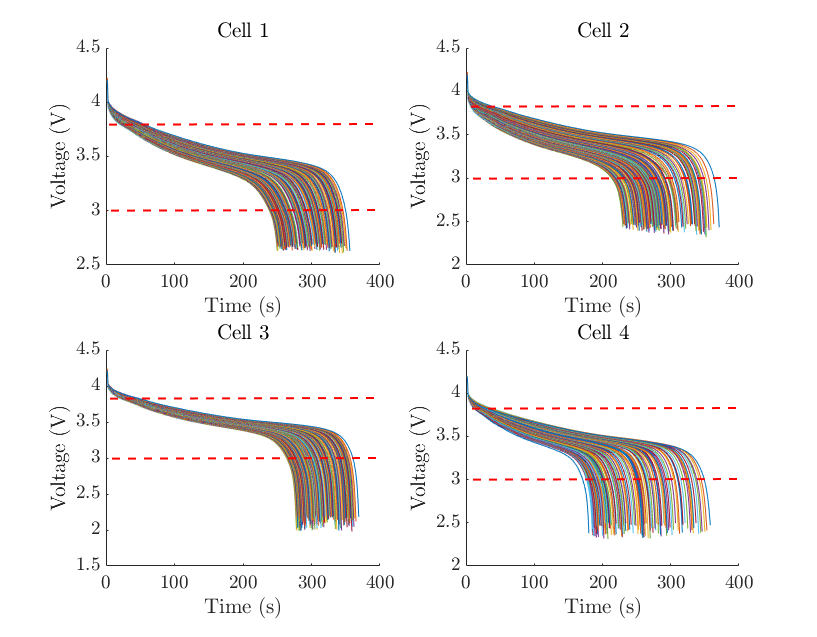}
    \caption{NASA dataset cell voltage profiles during cycling. Red lines demarcate the voltage window employed for subsequent analysis.}
    \label{ds_fig:3}
\end{figure}

\begin{figure}[!h]
    \centering
    \includegraphics[width=0.5\columnwidth]{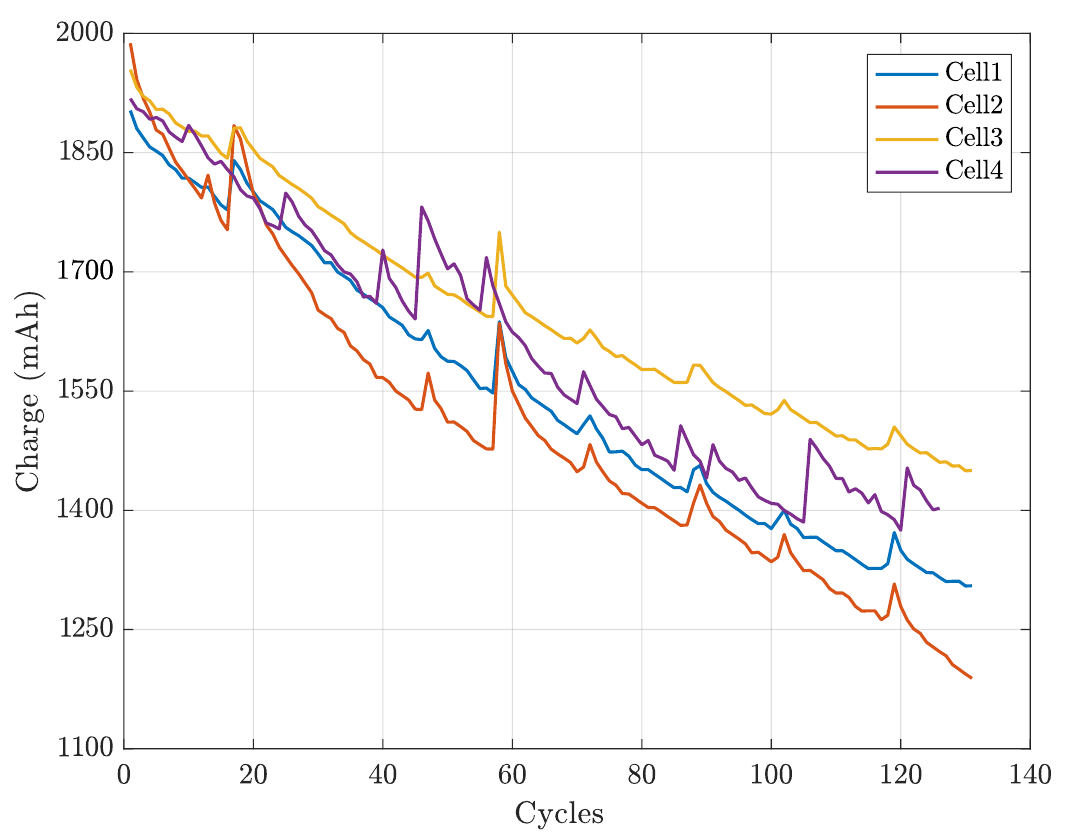}
    \caption{NASA dataset discharge capacity per cycle. Spikes indicate cell rest periods between successive cycles.}
    \label{ds_fig:4}
\end{figure}

Additionally, the NASA dataset \cite{NASA} was utilized. Four lithium-ion battery cells were used for this dataset. The upper voltage limit was 4.2 V while the lower voltage limit varied from 2.2 V to 2.7 V among the cells. \autoref{ds_fig:3} depicts the voltage profiles of the NASA cells. Unlike the Oxford dataset, the NASA voltage profiles exhibit greater variability across cycles \cite{ELEFTHERIADIS2024110889}.
\autoref{ds_fig:4} illustrates discharge capacity per cycle, demonstrating consistent degradation across all cells.
\subsection{PoliMi}
There are limited publicly accessible datasets available for researchers to validate their algorithms for SOC and SOH estimation. While there are sufficient datasets for SOC estimation, the availability is notably less for SOH, and most existing datasets do not include aging under dynamic driving cycles to mimic real-life conditions. For this reason, a series of experiments were designed and conducted in the battery laboratory of T.U. Berlin, with the collaboration of the Politecnico di Milano. The data associatted with the dataset can be found in Mendeley Data \cite{Eleftheriadis2024-sp}. This dataset comprises six NMC lithium-ion batteries. Cycling was performed at a controlled temperature of 35°C, with discharge profiles based on three driving simulation cycles. \autoref{ds_fig:PoliMi_fig3} clearly shows the degradation mechanism as a function of cycle number.
\begin{figure*}[ht!]
    \centering
    \includegraphics[width=0.8\textwidth]{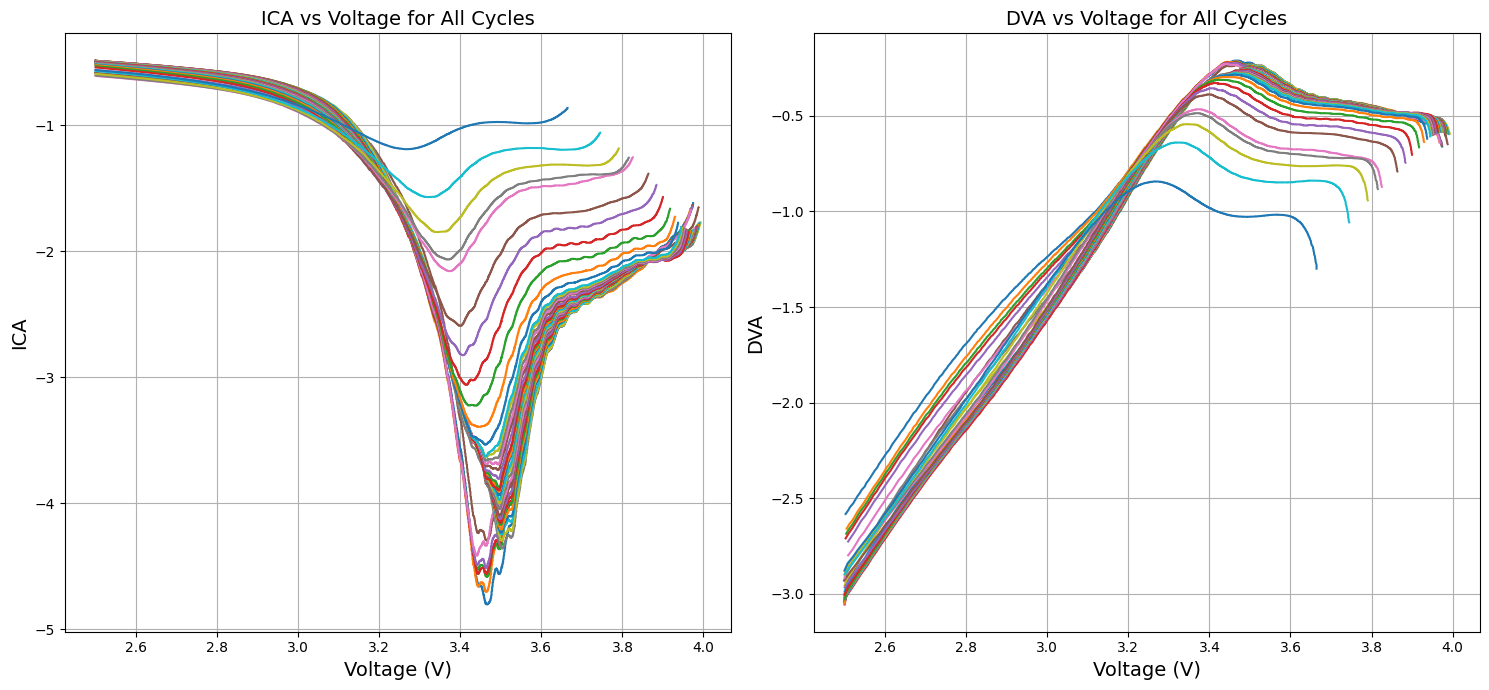}
    \caption{The ICA and DVA vs Voltage for all cycles for cell 1 of the PoliMi dataset. As the cell degrades, there is a noticeable trend of decreasing ICA/DVA values and changes in voltage.}
    \label{ds_fig:PoliMi_VQ}
\end{figure*}
\begin{figure}[!h]
    \centering
    \includegraphics[width=0.5\columnwidth]{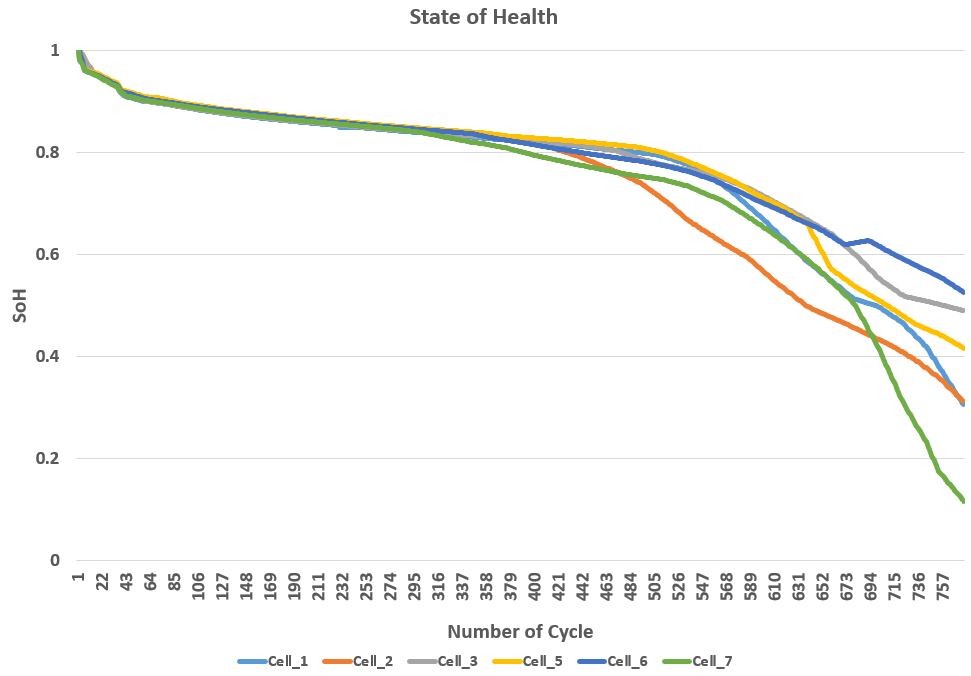}
    \caption{The evolution of the SOH of the NMC battery cells.}
    \label{ds_fig:PoliMi_fig3}
\end{figure}
\subsection{Input Feature Selection}
\begin{figure}[!hbtp]
    \centering
    \includegraphics[width=0.7\columnwidth]{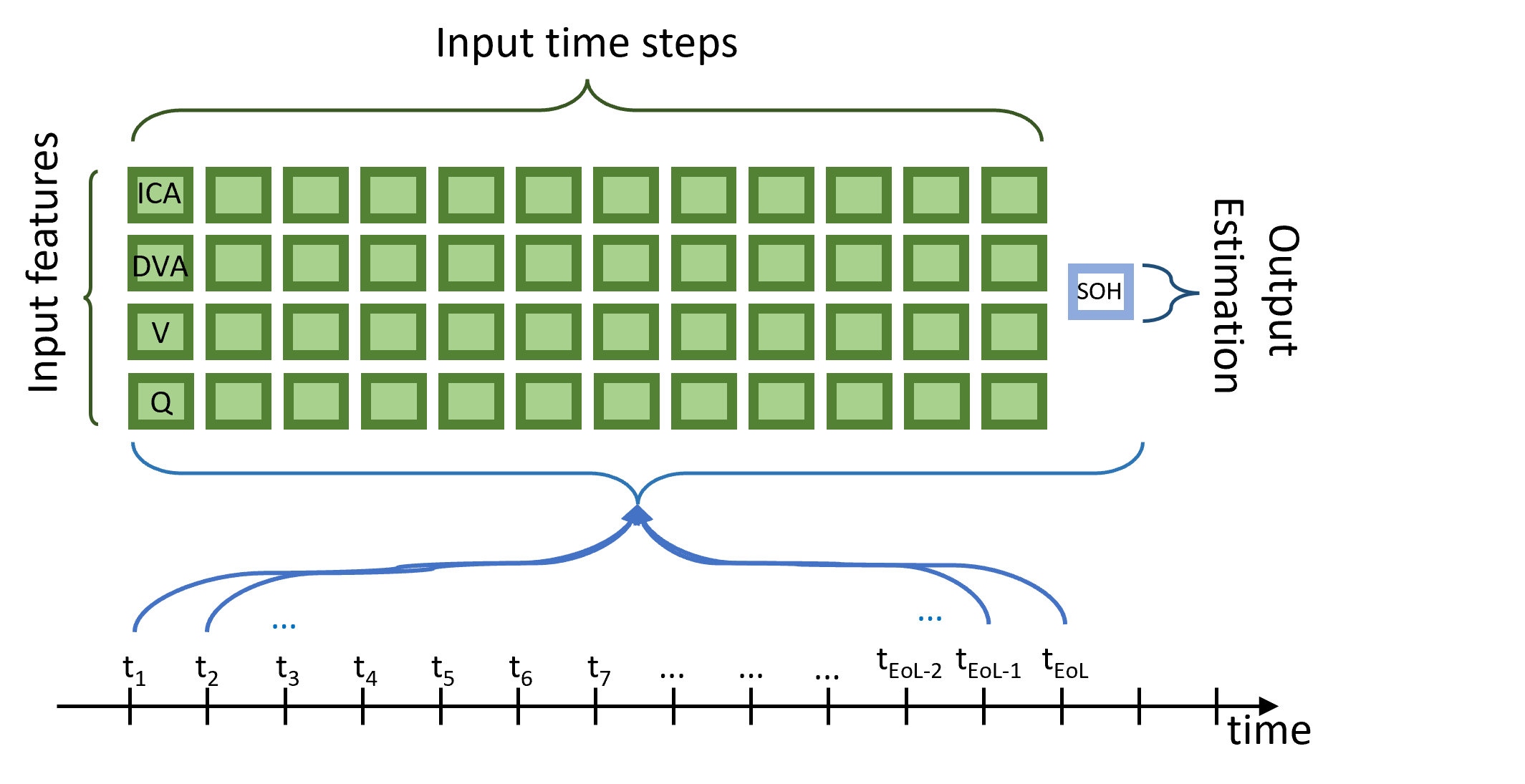}
    \caption{Time-series input shape and output estimation of each partial discharge cycle. At each time step t, an estimation is made, with the total number of steps being equal to the End of Life (EoL) of the cell, which occurs when the SOH reaches 70\%.}
    \label{soh:fig:TimeseriesArchitecture}
\end{figure}

\begin{figure}[!hbtp]
    \centering
    \includegraphics[width=0.65\columnwidth]{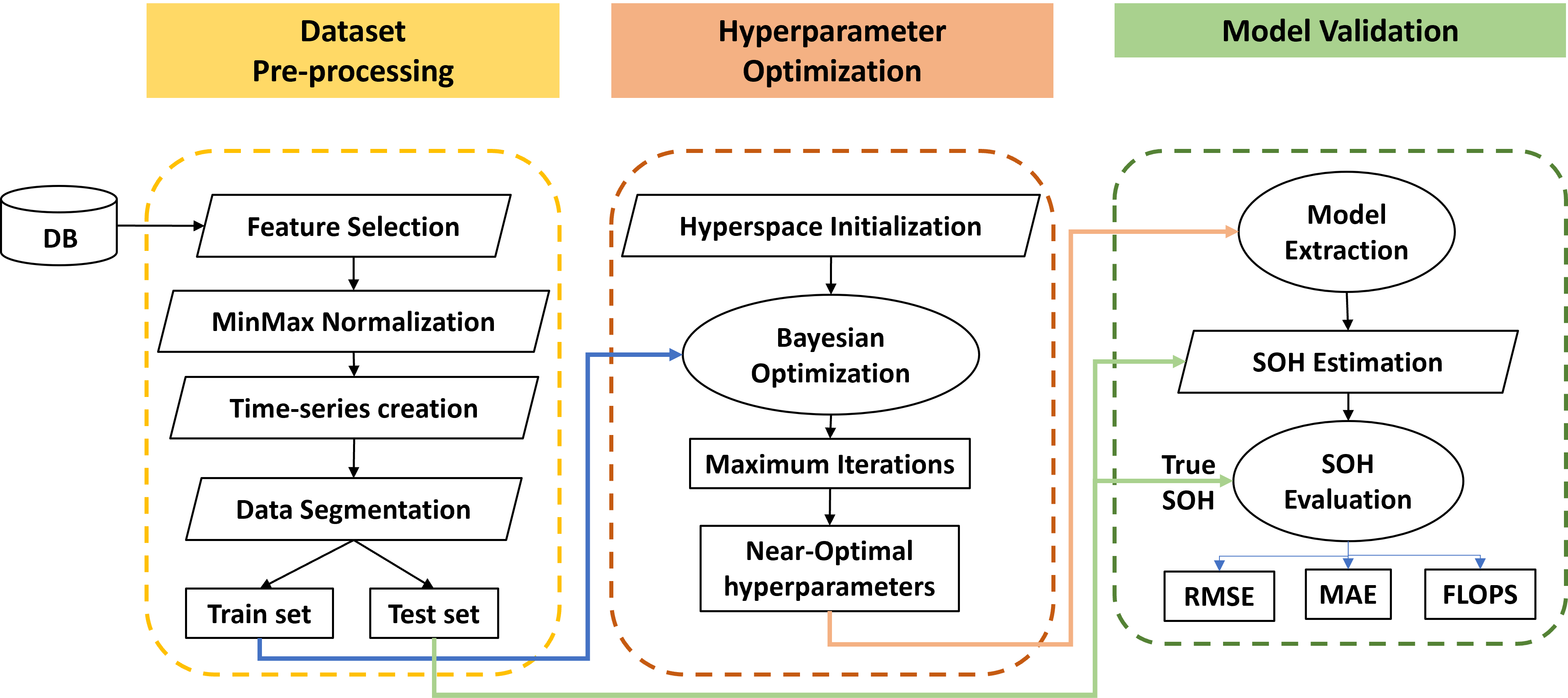}
    \caption{General framework of the proposed solution.}
    \label{soh:fig:TotalFramework}
\end{figure}

Unlike method described in \cite{ELEFTHERIADIS2024110889}, which extracted only peak ICA values, corresponding voltage points, and curve integration, this research employs four parameters as input features for the network: ICA, DVA, V, and Q from the discharging cycles. The complete dataset for these parameters will be leveraged to fully harness the capabilities of the RNN. The ICA and DVA values are presented for the PoliMi dataset as a refernce in \autoref{ds_fig:PoliMi_VQ}. For the other two datasets a similar behavior is observed.
Therefore, the input features will be the time-series representations of these parameters, creating time-series data for the entire ICA, DVA, V, and Q curves during partial discharging represented as [$ICA$, $DVA$, $V$, $Q$]$_{WS}$, where $WS$ denotes the window length. \autoref{soh:fig:TimeseriesArchitecture} depicts an example of the time-series input created for SOH estimation at each cycle.
For the Oxford and NASA datasets, the partial discharging voltage range will remain the same, between 3.4 to 4.1 V. In contrast, for the TUB dataset, the voltage range is configured between 3.1 and 3.7 V to capture all the information of the ICA peaks. Furthermore, to optimize the network's robustness, the sampling interval for the Oxford and PoliMi datasets was manually adjusted to a 10-second resolution, with no significant impact on the predictive results observed. Consequently, the WS of the input time series for the NASA, PoliMi, and Oxford datasets was set at 151, 117, and 238 time steps, respectively. For model training, cells, 1,2 and 4 were used for NASA, cells 1-8 for Oxford, and cell 1 for PoliMi. Model evaluation employed cells 3,7 and 8 from the PoliMi, NASA, and Oxford University datasets, respectively.
In the data preprocessing phase, the input dataset undergoes min-max scaling \cite{scikit-learn}. Through this transformation the data values range from 0 to 1, facilitating the neural network's ability to handle diverse input features effectively \cite{10407877}. The NN implementation is incorporated through the Keras Library \cite{chollet2015keras}. Additionally, the Bayesian Optimization framework, responsible for fine-tuning the model's hyperparameters, is skillfully integrated through the use of the Keras Tuner Library \cite{omalley2019kerastuner}. The Adam optimizer \cite{Kingma2014-tk} was used along with the searchable hyperparameter of the learning rate, while the Huber loss was chosen \cite{Huber1992-ys}. The network underwent training for 1000 epochs, incorporating an early stopping strategy if there was no improvement in validation loss after 50 epochs.
\autoref{soh:fig:TotalFramework} illustrates the overall structure of the proposed solution, which is segmented into three main areas: dataset preprocessing, hyperparameter optimization, and model validation. A batch size of 128 was employed, and a total of 50 trials were conducted to determine a set of hyperparameters that were close to optimal.
\section{Proposed Method}
\subsection{Convolutional Neural Network}
\begin{figure}[hbtp]
    \centering
    \includegraphics[width=0.55\textwidth]{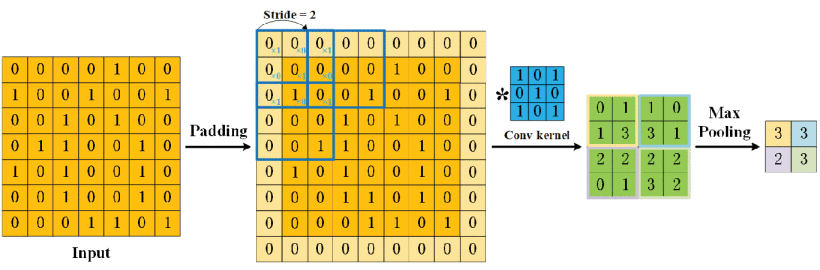}
    \caption{Convolutional Neural Network architecture \cite{9451544}.}
    \label{dd_fig:cnn}
\end{figure}

The CNN is one of the most prominent networks in the field of deep learning \cite{9451544}. The CNN is a type of Feed Forward Neural Network (FFNN) designed to extract features from data using convolution structures. Key features of the CNN include the establishment of local connections where each neuron is linked to only a few preceding neurons instead of all. Additionally, weight sharing in CNNs reduces the number of parameters needed. Finally, dimension reduction is achieved through pooling techniques, further decreasing the required number of parameters.
In constructing a CNN, four essential components are needed. The first is convolution, crucial for extracting features, resulting in feature maps. To address information loss at the borders during convolution, padding is used, which involves enlarging the input with zeros to indirectly modify its size. The stride, determining the convolution's density, is also important; a larger stride results in lower density. Finally, to prevent overfitting due to the abundance of features in the feature maps, pooling (or downsampling) is used to reduce redundancy. Pooling can be of two types: max pooling and average pooling. The structure of a typical CNN is presented in \autoref{dd_fig:cnn}.
The most typical types of convolutional layers are the `Conv1D', `Conv2D', and `Conv3D', each one dealing with data of different shapes. The `Conv1D' typically used for time-series data and sequential data applies a 1D convolutional operation with the data having a 2D or 3D dimension, with shape [`number of samples', `sequence length', `number of input features']. Furthermore, the `Conv2D' typically used for image data of data with spatial relationships applies a 2D convolutional operation with the data having usually 4D dimensions with shape [`number of samples', `height', `width', `channels']. Finally, the `Conv3D' is used for 3D data as volumetric data or videos and applies a 3D convolutional operation with the data having a 5D dimension with shape [`number of samples', `depth', `height', `width', `channels'].
\subsection{Bidirectional Long Short-Term Memory Neural Network}
\begin{figure}[hbtp]
    \centering
    \includegraphics[width=0.4\textwidth]{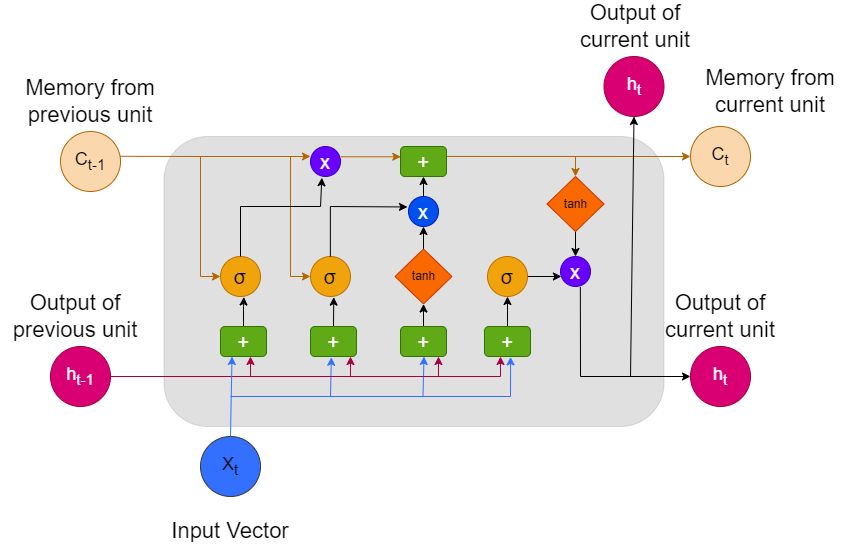}
    \caption{LSTM cell architecture \cite{BObilstm}}
    \label{fig:fig01}
\end{figure}
This research also employs LSTM neural networks. This type of RNN overcomes the vanishing gradients challenge by using gating mechanisms to control information flow. These gates include the input, output and forget gates. Analytical information about the aforementioned gates and their operation can be found in \cite{BObilstm}. As a result, they capture long-term dependencies effectively across data. The LSTM cell architecture is depicted in \autoref{fig:fig01}.
In summary, LSTMs' ability to retain relevant information while filtering noise makes them well-suited for sequential data processing.
Building on this, BiLSTMs process information bidirectionally, capturing information in both temporal directions. This results in more comprehensive feature representations compared to standard LSTM networks \cite{BObilstm}.
\begin{figure}[hbtp]
    \centering
    \includegraphics[width=0.4\textwidth]{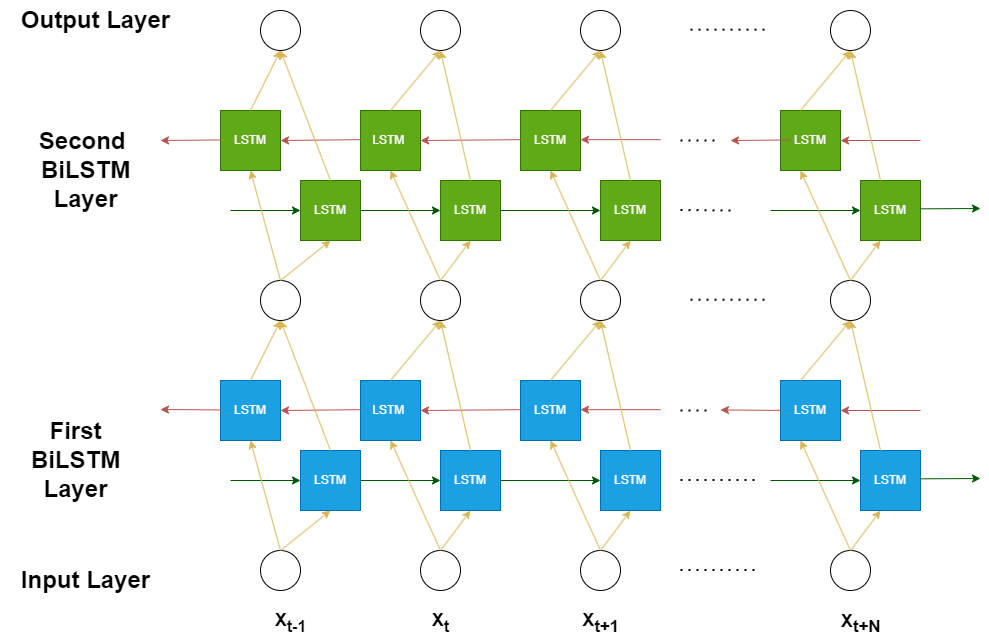}
    \caption{BiLSTM cell architecture \cite{10194767}}
    \label{fig:fig02}
\end{figure}
\autoref{fig:fig02} depicts the BiLSTM cell architecture. Information flows bidirectionally through forward and backward layers, enabling the network to capture temporal information from both past and future contexts \cite{BObilstm}.
LSTM networks with stacked layers excel at modeling battery behavior by capturing temporal dynamics and long-term dependencies in time-series measurements. In \cite{7862927}, the LSTM showcases remarkable performance in predicting SOH for a single time step, demonstrating its ability to process sequential data including voltage, current, and temperature measurements. Similarly, researchers in \cite{CHENG2021121022} implement a backpropagation LSTM for battery SOH prediction, validating the applicability and efficiency of such networks.
\begin{figure}[ht!]
    \centering
    \includegraphics[width=0.5\textwidth]{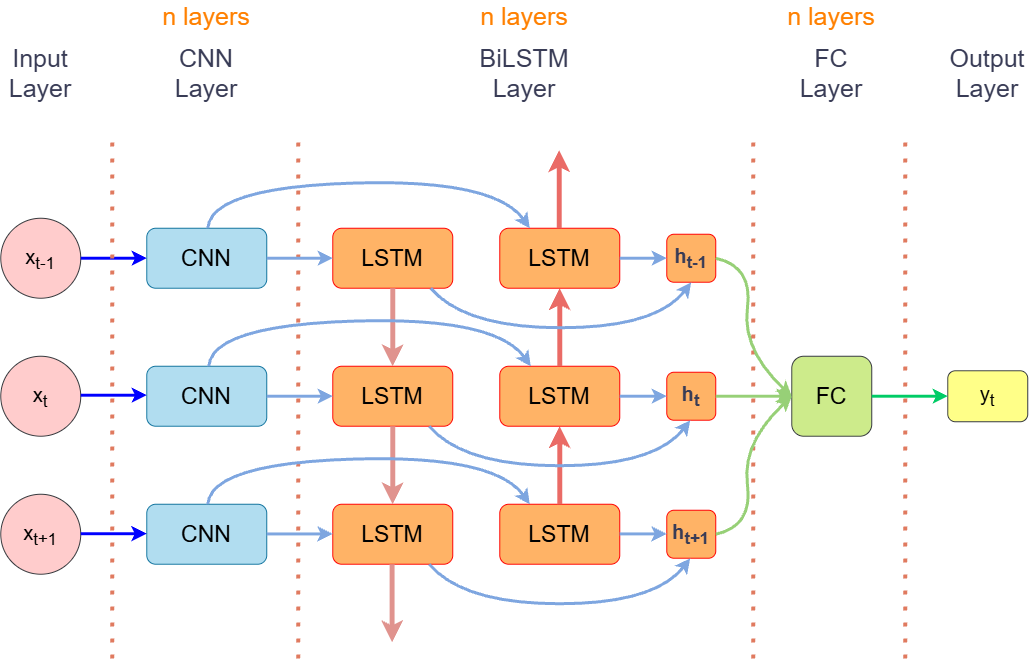}
    \caption{General Stacked CNN-BiLSTM architecture}
    \label{StackedBiLSTM}
\end{figure}
\subsection{Optimization Algorithm}
The main focus of this work lies on the use of Bayesian optimization in order to achieve optimal model hyperparameters. Compared to other exhaustive approaches such as grid and random search, BO demonstrates superiority in both efficiency and accuracy in high-dimensional space navigation \cite{frazier2018tutorial}. The representation of the objective function through stochastic models makes it possible to update its beliefs, thereby allowing it to focus on promising regions.
Bayesian optimization (BO) starts by establishing a prior distribution over the hyperparameters. In this way, it leverages all available prior information and evaluations regarding their behavior. At each iteration, the algorithm selects promising configurations taking into account previous performance results. Through this refinement process, BO navigates efficiently through the hyperparameter space, focusing on promising candidates and discarding unlikely solutions \cite{YANG2020295}. This results to the reduction of the computational costs of hyperparameter tuning.
\subsection{Training Procedure}
The configuration of the proposed network includes a number of key parameters, meticulously fine-tuned to improve its performance. A unique hyperspace has been developed for each hyperparameter to aid in the identification of the most effective, or near-optimal, hyperparameter settings. The following parameters were made searchable for the CNN-RNN network's hyperparameter search space:
\begin{itemize}
    \item Learning rate of the optimizer between the values of 0.001, 0.0001 and 0.00001.
    \item Two CNN layers with 1 to 15 filters, individually searchable, while the rest of the hyperparameters are fixed with a kernel size of 3, activation function being the Relu.
    \item One to three initial RNN layers with 1 to 15 units, (1-unit increments), returning the full sequence.
    \item One RNN layer within the range of 1 to 15 units, in increments of 1 unit, without the whole sequence being returned.
    \item A single RNN layer with 1-15 units (1-unit increments), without sequence return.
    \item One to three additional FC layers.
    \item FC layer units: 1-15 neurons per layer (1-neuron increments).
\end{itemize}
The goal of defining the search space for hyperparameters is to facilitate a comprehensive probabilistic exploration through Bayesian optimization, ensuring that no possible configuration is missed. This detailed strategy allows for the precise adjustment of the proposed network, enhancing its predictive accuracy and robustness, thereby aligning with the study's objectives for more precise and reliable predictions.
To enhance accuracy and robustness, multiple stacked layers were added to every stage of the final algorithm, namely the CNN, RNN and FC layers. \autoref{StackedBiLSTM} depicts the general framework of the proposed solution. Specifically, $x$ represents the sequential data input previously discussed, and $y_t$ denotes the SOH estimation output. The figure illustrates the $n$ stacked layers at each intermediate stage of the model, which substantially improve performance.
\section{Results}
\begin{figure}[h]
    \centering
    \includegraphics[width=0.5\columnwidth]{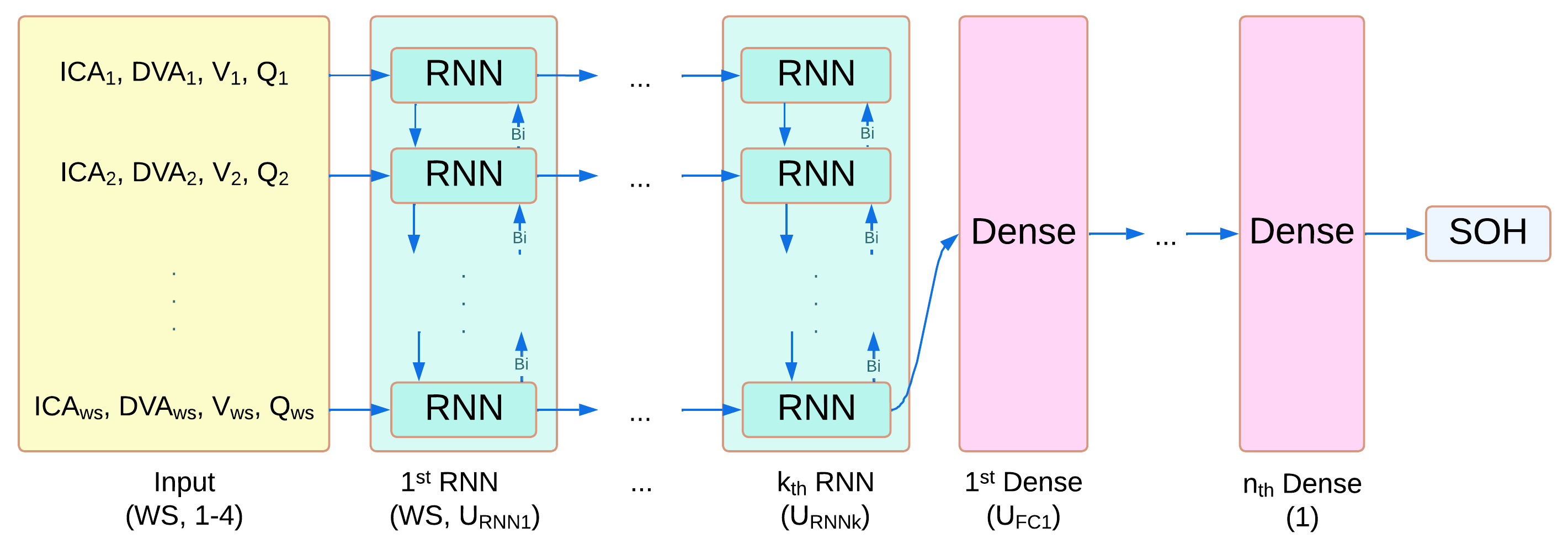}
    \caption{Second category of stacked RNN layers.}
    \label{soh_tl:2nd_category}
\end{figure}

\begin{figure}[h]
    \centering
    \includegraphics[width=0.5\columnwidth]{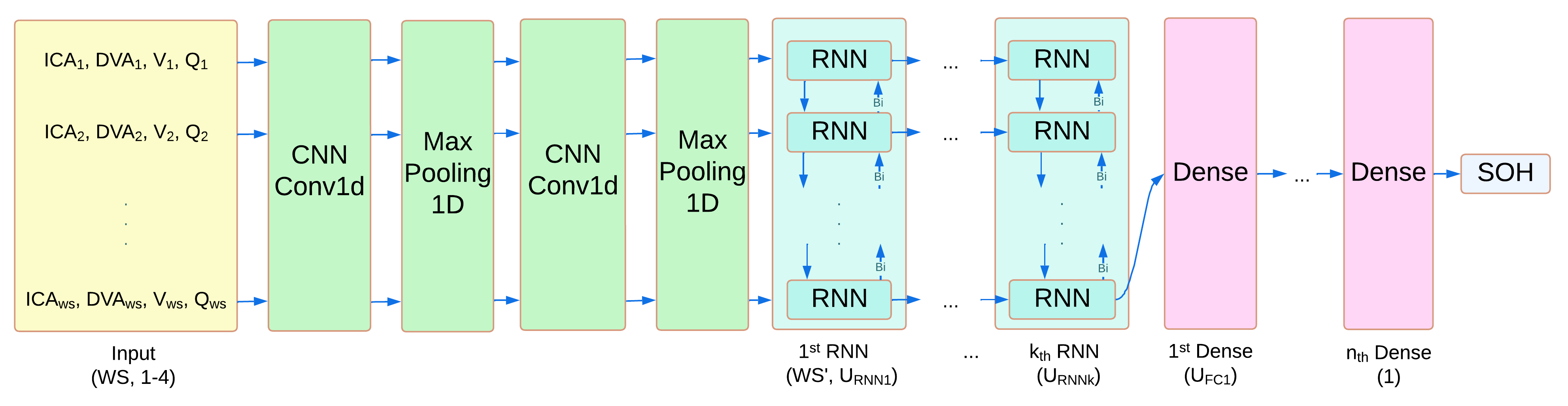}
    \caption{Third category of the combination of CNN and RNN layers.}
    \label{soh_tl:3rd_category}
\end{figure}

\begin{figure}[h]
    \centering
    \includegraphics[width=0.5\linewidth]{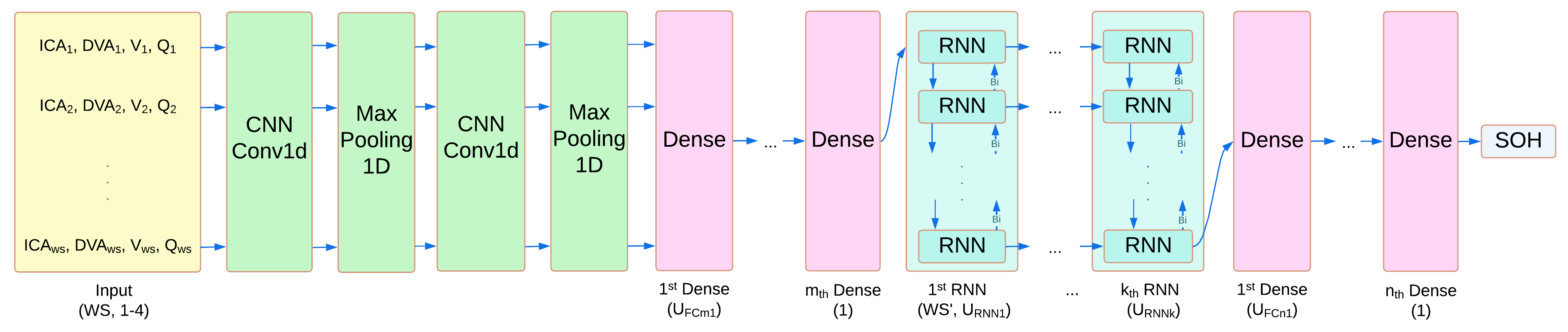}
    \caption{Fourth category of the combination of CNN, FC, and RNN layers.}
    \label{soh_tl:4th_category}
\end{figure}

The prediction performance of the proposed model was assessed using the RMSE and MAE metrics. The definitions of these two error metrics are given by \autoref{eqn:rmse} and \autoref{eqn:mae}, respectively. As the SOH is a percentage, the corresponding error metric will be reported in percentage (\%).
\begin{equation}
\label{eqn:rmse}
     \mathit{RMSE} = \sqrt{\frac{1}{n}\sum_{i=1}^{n}{(SOH_i - \hat{SOH_i})^2}}
\end{equation}
\begin{equation}
\label{eqn:mae}
    \mathit{MAE}=\frac{1}{n} \sum_{j=1}^n \lvert SOH_{i}-\hat{SOH_i}\rvert \\[2ex]
\end{equation}
Where $SOH_i$ and $\hat{SOH_i}$ refer to the battery's real and predicted SOH value, respectively.
Additionally, to assess the computational burden required for the SOH estimation, the total number of FLOPs is considered for every candidate solution. FLOPs represent the number of floating-point operations needed to run a single instance of the given model, the calculation of which was performed using the Keras library \cite{FLOPs}.
For each dataset, the BOGP algorithm was applied to fine-tune a total of thirteen different architectures within the designated hyperspace for each of the five input feature combinations. The evaluation metrics were then used for comparison. These thirteen architectures are detailed in \autoref{soh:models}. The first is the baseline network, a simple FFNN. The second category comprises straightforward stacked RNNS, the LSTM, the BiLSTM, the GRU, and the BiGRU NNs depicted in \autoref{soh_tl:2nd_category}. The next category involves combinations of stacked CNN and stacked RNN with the previously mentioned layers, presented in \autoref{soh_tl:3rd_category}. The last category is similar to the third, but with the addition of FC layers between the CNN and RNN layers, presented in \autoref{soh_tl:4th_category}, representing the proposed method.
\begin{table}[h]
\centering
\caption{Models employed in the comparison to find the one yielding the best results.}
\label{soh:models}
\begin{adjustbox}{width=0.8\columnwidth}
\begin{tabular}{c|cccc}
\hline
Category & \multicolumn{4}{c}{Models} \\ \hline\hline
FNN & \multicolumn{4}{c}{MLP} \\ \hline
RNN & \multicolumn{1}{c|}{LSTM} & \multicolumn{1}{c|}{BiLSTM} & \multicolumn{1}{c|}{GRU} & BiGRU \\ \hline
CNN-RNN & \multicolumn{1}{c|}{CNN-LSTM} & \multicolumn{1}{c|}{CNN-BiLSTM} & \multicolumn{1}{c|}{CNN-GRU} & CNN-BiGRU \\ \hline
CNN-FC-RNN & \multicolumn{1}{c|}{CNN-FC-LSTM} & \multicolumn{1}{c|}{CNN-FC-BiLSTM} & \multicolumn{1}{c|}{CNN-FC-GRU} & CNN-FC-BiGRU \\
\end{tabular}
\end{adjustbox}
\end{table}
\subsection{Results on PoliMi dataset}
This section evaluates the performance of the proposed model on the PoliMi dataset. The results of the best three models are presented in \autoref{soh:polimi}. The highest predictive performance was achieved with a model of the last category, where the CNN was integrated in a cascading manner with FC layers, followed by RNN layers. In this setup, the most effective RNN layer proved to be the BiLSTM layer. This configuration, particularly when using ICA and voltage as inputs, yielded a MAE of 0.12\% and a RMSE of 0.14\%. The comparison between the estimated values and the actual values can be viewed in \autoref{soh:polimi_est}. The rest of the results of the exhaustive analysis can be found in \autoref{appendixA}.
\begin{figure}[h]
    \centering
    \includegraphics[width=0.55\columnwidth]{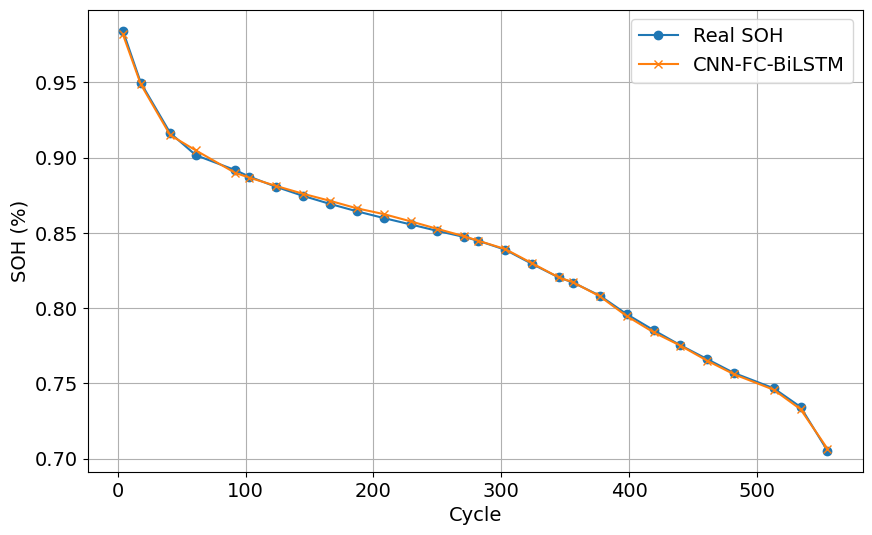}
    \caption{Estimated values of SOH for the PoliMi dataset of the CNN-FC-BiLSTM model with inputs the ICA, DVA and the voltage.}
    \label{soh:polimi_est}
\end{figure}
\begin{table}[h]
\centering
\caption{The top three models for the PoliMi dataset achieving the lowest errors with the corresponding input features, MAE, RMSE of the estimated SOH and the FLOPS}
\label{soh:polimi}
\begin{adjustbox}{width=0.6\columnwidth}
\begin{tabular}{c|ccc}
Model & CNN+FC+BiLSTM & CNN+FC+LSTM & BiGRU \\ \hline
Input & ICA,  DVA, V & ICA & ICA, V, Q \\ \hline
MAE & 0.12 & 0.12 & 0.16 \\
RMSE & 0.14 & 0.15 & 0.18 \\
FLOPS & 6424192 & 8605504 & 60544
\end{tabular}%
\end{adjustbox}
\end{table}
This solution entails a high number of FLOPS, making it computationally intensive. It would be beneficial to be implemented in a cloud-based environment to avoid the need for additional computational resources in the BMS of the EV.
If computational constraints limit the application, the results based on FLOPS with the smallest size of the developed model are presented in \autoref{soh:polimi_flops}. Combining the smallest size with the best predictive capabilities among the three cases presented in \autoref{soh:polimi_flops}, the first case is the most favourable of the GRU as a model with FLOPS of 2176, achieving a MAE of 0.17\% and a RMSE of 0.20\%.
\begin{table}[h]
\centering
\caption{The top three models for the PoliMi dataset achieving the minimum size with the corresponding input features, MAE, RMSE of the estimated SOH and the FLOPS}
\label{soh:polimi_flops}
\begin{adjustbox}{width=0.5\columnwidth}
\begin{tabular}{c|ccc}
Model & GRU & LSTM & GRU \\ \hline
Input & ICA, V & ICA, DVA, V, Q & ICA, DVA, V, Q \\ \hline
MAE & 0.17 & 0.20 & 0.31 \\
RMSE & 0.20 & 0.23 & 0.40 \\
FLOPS & 2176 & 2176 & 2176
\end{tabular}%
\end{adjustbox}
\end{table}
The minimal computational cost of this configuration, based on the minimum FLOPS, coupled with its acceptable accuracy, offer a practical approach for online applications, implementing the model into a microcontroller of a BMS.
\subsection{Results on Oxford dataset}
This section focuses on analyzing the performance of the proposed model using the Oxford dataset. The outcomes from the top three models are detailed in \autoref{soh:oxford}. Once more, the model from the last category, integrating a CNN in a cascading sequence with FC layers, followed by RNN layers, achieved the best predictive performance. Within this configuration, the BiLSTM layer emerged as the most effective RNN layer. This configuration, particularly when using ICA, DVA and voltage as inputs, yielded a MAE of 0.16\% and a RMSE of 0.19\%. The comparison between the estimated values and the actual values can be viewed in \autoref{soh:oxford_est}. The rest of the results of the exhaustive analysis can be found in \autoref{appendixA}.
\begin{figure}[h]
    \centering
    \includegraphics[width=0.55\columnwidth]{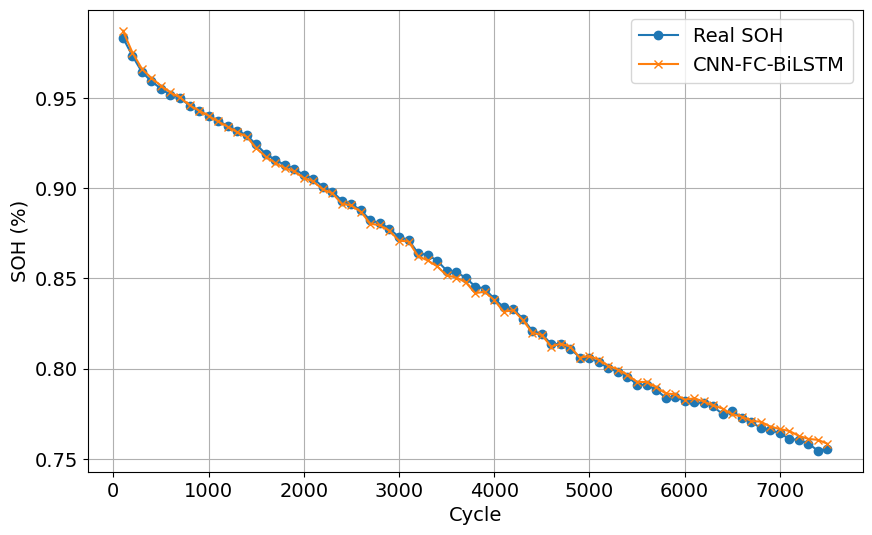}
    \caption{Estimated values of SOH for the Oxford dataset of the CNN-FC-BiLSTM model with inputs the ICA and the voltage.}
    \label{soh:oxford_est}
\end{figure}
\begin{table}[h]
\centering
\caption{The top three models for the Oxford dataset achieving the lowest errors with the corresponding input features, MAE, RMSE of the estimated SOH and the FLOPS}
\label{soh:oxford}
\begin{adjustbox}{width=0.6\columnwidth}
\begin{tabular}{c|ccc}
Model & CNN+FC+BiLSTM & CNN+FC+LSTM & CNN+BiGRU \\ \hline
Input & ICA, V & ICA, DVA, V, Q & ICA, V \\ \hline
MAE & 0.16 & 0.16 & 0.17 \\
RMSE & 0.19 & 0.20 & 0.23 \\
FLOPS & 15076288 & 9379712 & 1630080
\end{tabular}%
\end{adjustbox}
\end{table}
Similar to the previous dataset, this solution requires a substantial number of FLOPS, leading to high computational demands. Implementing it in a cloud-based setting would be advantageous, while if computational limitations are once again a concern, the outcomes based on FLOPS for the most compact version of the developed model are shown in \autoref{soh:oxford_flops}. Among the three scenarios presented there, the most favorable is the final case with the LSTM as model. It strikes a balance between size and performance, with FLOPS of 2305 and achieving an MAE of 0.22\% and an RMSE of 0.26\%.
\begin{table}[h]
\centering
\caption{The top three models for the Oxford dataset achieving the minimum size with the corresponding input features, MAE, RMSE of the estimated SOH and the FLOPS}
\label{soh:oxford_flops}
\begin{adjustbox}{width=0.5\columnwidth}
\begin{tabular}{c|ccc}
Model & LSTM & LSTM & LSTM \\ \hline
Input & ICA, DVA, V, Q & ICA, V & ICA, DVA, V \\ \hline
MAE & 0.25 & 0.44 & 0.22 \\
RMSE & 0.29 & 0.54 & 0.26 \\
FLOPS & 2176 & 2176 & 2304
\end{tabular}%
\end{adjustbox}
\end{table}
The reduced computational burden of this approach based on the minimum FLOPS, coupled with its acceptable accuracy, demonstrats the feasibility of implementing the model into a microcontroller of a BMS.
\subsection{Results on NASA dataset}
This final section is dedicated to evaluating the performance of the proposed model with the NASA dataset. Detailed results of the top three models are presented in \autoref{soh:nasa}. This time, the model from the third category, which combines a CNN in a cascading connection with the RNN layers, without the intermediate connection of FC layers, delivered the highest predictive accuracy. The BiGRU layer within this setup, this time, proved to be the most effective RNN layer. When utilizing solely the ICA as input, this configuration achieved a MAE of 0.45\% and a RMSE of 0.71\%. The comparison between estimated and actual values is available in \autoref{soh:nasa_est}. Comprehensive results from the extensive analysis are available in \autoref{appendixA}.
\begin{figure}[h]
    \centering
    \includegraphics[width=0.55\columnwidth]{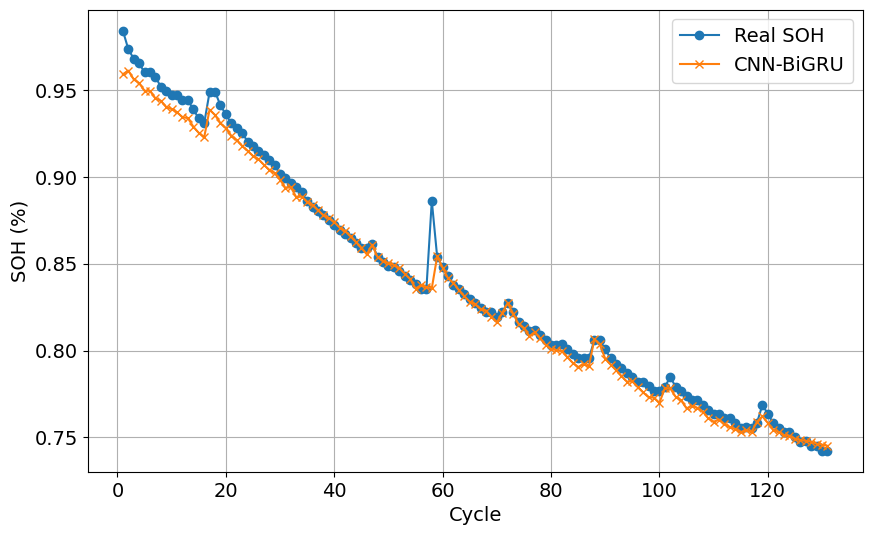}
    \caption{Estimated values of SOH for the NASA dataset of the CNN-BiGRU model  with input the ICA}
    \label{soh:nasa_est}
\end{figure}
\begin{table}[h]
\centering
\caption{The top three models for the NASA dataset achieving the lowest errors with the corresponding input features, MAE, RMSE of the estimated SOH and the FLOPS}
\label{soh:nasa}
\begin{adjustbox}{width=0.4\columnwidth}
\begin{tabular}{c|ccc}
Model & CNN+BiGRU & BiGRU & FFNN \\ \hline
Input & ICA & ICA & ICA \\ \hline
MAE & 0.45 & 0.53 & 0.56 \\
RMSE & 0.71 & 0.72 & 0.81 \\
FLOPS & 5799424 & 4096 & 327104
\end{tabular}%
\end{adjustbox}
\end{table}
As with the previous dataset, this solution also involves a high number of FLOPS, necessitating significant computational resources. Opting for a cloud-based implementation would be beneficial in this context. However, if computational constraints are a concern, the results for the most compact model version, based on FLOPS, are displayed in \autoref{soh:nasa_flops}. Of the three scenarios outlined there, the most advantageous is the third case, which uses the BiGRU model while being the second-best model overall based on the error metrics as it can be seen in \autoref{soh:nasa}. This option offers a good compromise between size and efficacy, with FLOPS of 4096 and achieving a MAE of 0.53\% and a RMSE of 0.72\%.
\begin{table}[h]
\centering
\caption{The top three models for the NASA dataset achieving the minimum size with the corresponding input features, MAE, RMSE of the estimated SOH and the FLOPS}
\label{soh:nasa_flops}
\begin{adjustbox}{width=0.4\columnwidth}
\begin{tabular}{c|ccc}
Model & LSTM & LSTM & LSTM \\ \hline
Input & ICA, V & ICA, V, Q & ICA \\ \hline
MAE & 1.76 & 1.01 & 0.53 \\
RMSE & 2.23 & 1.17 & 0.72 \\
FLOPS & 512 & 640 & 4096
\end{tabular}%
\end{adjustbox}
\end{table}
The reduced computational burden of this approach, based on the minimum FLOPS, coupled with its acceptable accuracy, provide a viable approach for real-time operational scenarios, implementing the model into a microcontroller of a BMS.
\section{Discussion}
The experimental results confirm the effectiveness of the proposed hybrid deep learning framework of combination of CNN and RNN moddels for accurate SOH estimation across three publicly available datasets. Specifically, the CNN-FC-BiLSTM model yielded the best performance on the Oxford and Polimi datasets, while the CNN-BiGRU model achieved superior results on the NASA dataset.
For the Oxford and Polimi datasets, the CNN-FC-BiLSTM model successfully leveraged the strengths of convolutional layers for spatial feature extraction, fully connected layers for nonlinear transformation, and BiLSTM layers for capturing bidirectional temporal dependencies in the degradation patterns. The consistent performance across different battery cells within these datasets demonstrates the model's ability to generalize well under relatively stable and homogeneous operating conditions.
In contrast, the NASA dataset—characterized by higher noise levels and more aggressive degradation—was better handled by the CNN-BiGRU model. GRU units, known for their simpler structure compared to LSTMs, are often more efficient and robust in the presence of noisy or sparse data \cite{Jzefowicz2015AnEE}. The superior performance of the CNN-BiGRU model in this context suggests that GRUs may offer advantages in real-world BMS applications where data quality is not always ideal.
These results also highlight the importance of model selection and architecture design in battery health diagnostics. A one-size-fits-all model may not be optimal across diverse datasets, and hybrid architectures that integrate convolutional and recurrent elements offer a flexible foundation for adapting to different battery types and usage profiles.
An additional observation is the benefit of combining multiple input features, such as voltage, current, capacity, and more complex post-processing of those to enrich the model's understanding of battery behavior. This supports the use of multi-modal time-series data in SOH estimation, encouraging further research into data fusion techniques and sensor integration.
Nevertheless, challenges remain. The current models assume fixed sequence lengths. Future work could investigate dynamic input handling with different input sequences for greater deployment flexibility.
In conclusion, the presented results validate the hybrid CNN-based architectures as powerful tools for SOH estimation. These insights can inform the development of next-generation battery management systems that are both accurate and adaptable.
\section{Conclusions}
This research presents a novel approach for battery SOH estimation combining cascaded CNN and RNN layers. For PoliMi and Oxford datasets, BiLSTM layers with intermediate FC connections provided optimal performance, while the NASA dataset benefited from BiGRU layers without FC layers. Bayesian Optimization enabled systematic hyperparameter tuning, reducing prediction errors and improving generalization. Comparative analysis demonstrated superior accuracy relative to existing machine learning techniques. Performance validation across multiple datasets confirms the method's effectiveness in capturing battery degradation dynamics. These results highlight the practical value of this framework for real-time battery management systems requiring accurate SOH assessment.
Model architecture selection proves critical in optimizing SOH estimation accuracy across heterogeneous datasets. CNN-FC-BiLSTM demonstrated superior performance on the PoliMi and Oxford datasets, while CNN-BiGRU excels under the diverse conditions of the NASA dataset. This underscores that architectural configuration selection must be informed by dataset-specific degradation patterns. Bayesian hyperparameter tuning and integration of multi-modal features enhanced the robustness and generalizability of the framework, rendering it suitable for deployment across various operational environments.
Future research could focus on exploring the advantages of adding attention mechanisms or transformer-based architectures to enhance the model's ability to capture long-term dependencies in battery behavior. Additionally, investigating transfer learning approaches to utilize insights from similar battery datasets for refining the models for specific scenarios or battery chemistries is a promising direction. Conducting comprehensive experiments with a variety of battery datasets to assess the model's ability to generalize across different battery types and conditions is also essential. Lastly, for practical implementation and to reduce computational expenses, methods such as model pruning and weight quantization could be considered to reduce the model's size.

\bibliography{mybib}
\bibliographystyle{IEEEtran}

\appendices
\section{Tables}
\label{appendixA}
\begin{table}[!h]
\centering
\caption{Errors and FLOPS for the first category of FFNN}
\begin{adjustbox}{width=1\columnwidth}
\begin{tabular}{|c|ccc|ccc|ccc|} \hline
Dataset & \multicolumn{3}{c|}{Oxford} & \multicolumn{3}{c|}{NASA} & \multicolumn{3}{c|}{PoliMi} \\ \hline
Features & MAE (\%) & RMSE (\%) & FLOPs (k) & MAE (\%) & RMSE (\%) & FLOPS (k) & MAE (\%) & RMSE (\%) & FLOPS (k) \\ \hline
ICA & 0.18 & 0.22 & 70016 & 0.56 & 0.81 & 327104 & 0.58 & 0.71 & 357248 \\
ICA, V & 0.23 & 0.30 & 155136 & 2.09 & 2.21 & 456320 & 0.30 & 0.40 & 582784 \\
ICA, V, Q & 0.33 & 0.38 & 126080 & 2.25 & 2.31 & 795904 & 0.26 & 0.32 & 991744 \\
ICA, DVA, V & 0.21 & 0.27 & 1306240 & 3.02 & 3.08 & 737024 & 0.27 & 0.40 & 872704 \\
ICA, DVA, V,   Q & 0.25 & 0.30 & 1922496 & 0.88 & 1.15 & 60160 & 0.49 & 0.52 & 1248832\\  \hline
\end{tabular}
\end{adjustbox}
\label{SOH:rnn}
\end{table}
\begin{table}[!h]
\centering
\caption{Errors and FLOPS for the second category of RNNs}
\begin{adjustbox}{width=1\columnwidth}
\begin{tabular}{|c|c|ccc|ccc|ccc|} \hline
\multicolumn{2}{|r}{Dataset} & \multicolumn{3}{c|}{Oxford} & \multicolumn{3}{c|}{NASA} & \multicolumn{3}{c|}{PoliMi} \\ \hline
Layer & Features & MAE (\%) & RMSE (\%) & FLOPs (k) & MAE (\%) & RMSE (\%) & FLOPS (k) & MAE (\%) & RMSE (\%) & FLOPS (k) \\ \hline
\multirow{5}{*}{LSTM}
 & ICA & 0.49 & 0.60 & 64.4 & 0.63 & 0.94 & 48.1 & 0.23 & 0.26 & 91.3 \\
 & ICA, V & 0.44 & 0.54 & 2.2 & 0.93 & 1.30 & 6.8 & 0.32 & 0.42 & 64.4 \\
 & ICA, V, Q & 0.24 & 0.31 & 4.9 & 1.01 & 1.17 & 0.6 & 0.24 & 0.27 & 9.0 \\
 & ICA, DVA, V & 0.22 & 0.26 & 2.3 & 0.85 & 1.09 & 65.9 & 0.55 & 0.61 & 31.7 \\
 & ICA, DVA, V, Q & 0.25 & 0.29 & 2.2 & 2.89 & 3.07 & 36.6 & 0.20 & 0.23 & 2.2 \\ \hline
\multirow{5}{*}{BiLSTM}
 & ICA & 0.34 & 0.38 & 60.5 & 0.88 & 1.25 & 7.7 & 0.44 & 0.52 & 60.5 \\
 & ICA, V & 0.35 & 0.39 & 4.5 & 1.76 & 2.23 & 0.5 & 0.20 & 0.25 & 60.5 \\
 & ICA, V, Q & 0.61 & 0.72 & 7.7 & 1.86 & 2.05 & 38.5 & 0.21 & 0.24 & 45.2 \\
 & ICA, DVA, V & 0.65 & 0.76 & 120.1 & 1.01 & 1.30 & 11.0 & 0.26 & 0.31 & 9.0 \\
 & ICA, DVA, V, Q & 0.26 & 0.34 & 4.5 & 1.42 & 1.72 & 60.9 & 0.19 & 0.26 & 59.0 \\ \hline
\multirow{5}{*}{GRU}
 & ICA & 0.89 & 1.08 & 2.6 & 2.81 & 2.95 & 6.8 & 0.45 & 0.56 & 7.0 \\
 & ICA, V & 0.80 & 0.96 & 4.6 & 1.56 & 1.80 & 6.4 & 0.17 & 0.20 & 2.2 \\
 & ICA, V, Q & 0.49 & 0.56 & 61.7 & 0.71 & 0.93 & 31.7 & 0.21 & 0.27 & 36.6 \\
 & ICA, DVA, V & 0.49 & 0.58 & 91.3 & 1.97 & 2.28 & 9.0 & 0.41 & 0.43 & 31.7 \\
 & ICA, DVA, V, Q & 0.41 & 0.46 & 9.3 & 2.05 & 2.39 & 61.5 & 0.31 & 0.40 & 2.2 \\ \hline
\multirow{5}{*}{BiGRU}
 & ICA & 0.34 & 0.45 & 4.1 & 0.53 & 0.72 & 4.1 & 0.32 & 0.41 & 6.8 \\
 & ICA, V & 0.50 & 0.58 & 66.3 & 1.38 & 1.56 & 4.1 & 0.69 & 0.77 & 36.5 \\
 & ICA, V, Q & 0.56 & 0.66 & 4.5 & 0.91 & 1.10 & 60.5 & 0.16 & 0.18 & 60.5 \\
 & ICA, DVA, V & 0.58 & 0.64 & 9.2 & 1.90 & 2.11 & 68.2 & 0.26 & 0.34 & 5.9 \\
 & ICA, DVA, V, Q & 0.91 & 1.05 & 11.6 & 0.65 & 0.91 & 54.4 & 0.84 & 0.88 & 8.8 \\  \hline
\end{tabular}
\end{adjustbox}
\label{SOH:rnn2}
\end{table}
\begin{table}[!h]
\centering
\caption{Errors and FLOPS for the third category of the combination of CNN and RNNs}
\begin{adjustbox}{width=1\columnwidth}
\begin{tabular}{|c|c|ccc|ccc|ccc|} \hline
\multicolumn{2}{|r}{Dataset} & \multicolumn{3}{c|}{Oxford} & \multicolumn{3}{c|}{NASA} & \multicolumn{3}{c|}{PoliMi} \\ \hline
Layer & Features & MAE (\%) & RMSE (\%) & FLOPs (k) & MAE (\%) & RMSE (\%) & FLOPS (k) & MAE (\%) & RMSE (\%) & FLOPS (k) \\  \hline
\multirow{5}{*}{CNN-LSTM}
 & ICA & 0.18 & 0.24 & 12111.5 & 0.77 & 0.98 & 5795.0 & 0.40 & 0.45 & 7025.0 \\
 & ICA, V & 0.33 & 0.37 & 13419.5 & 1.61 & 1.80 & 5281.3 & 0.21 & 0.26 & 1507.1 \\
 & ICA, V, Q & 0.19 & 0.22 & 14778.5 & 1.51 & 1.81 & 182.0 & 0.35 & 0.41 & 9223.7 \\
 & ICA, DVA, V & 0.26 & 0.33 & 14778.5 & 0.81 & 0.97 & 471.0 & 0.18 & 0.22 & 1984.0 \\
 & ICA, DVA, V, Q & 0.26 & 0.31 & 14089.0 & 1.43 & 1.52 & 5142.8 & 0.33 & 0.41 & 10113.7 \\  \hline
\multirow{5}{*}{CNN-BiLSTM}
 & ICA & 0.19 & 0.24 & 2499.6 & 0.72 & 0.97 & 613.2 & 0.25 & 0.29 & 3267.5 \\
 & ICA, V & 0.17 & 0.23 & 1630.1 & 0.91 & 1.23 & 2280.4 & 0.22 & 0.24 & 8421.2 \\
 & ICA, V, Q & 0.21 & 0.24 & 7074.9 & 1.45 & 1.68 & 7130.0 & 0.21 & 0.24 & 5856.8 \\
 & ICA, DVA, V & 0.23 & 0.27 & 14785.3 & 0.96 & 1.25 & 7074.3 & 0.30 & 0.37 & 5840.9 \\
 & ICA, DVA, V, Q & 0.31 & 0.36 & 16169.5 & 0.99 & 1.28 & 7755.9 & 0.65 & 0.78 & 10087.7 \\  \hline
\multirow{5}{*}{CNN-GRU}
 & ICA & 0.39 & 0.48 & 2586.8 & 0.70 & 1.03 & 5741.1 & 0.23 & 0.28 & 7534.1 \\
 & ICA, V & 0.40 & 0.49 & 8645.6 & 0.85 & 1.10 & 441.7 & 0.23 & 0.27 & 1749.2 \\
 & ICA, V, Q & 0.19 & 0.22 & 1987.2 & 1.09 & 1.22 & 7068.5 & 0.39 & 0.45 & 4407.6 \\
 & ICA, DVA, V & 0.32 & 0.41 & 1987.2 & 1.90 & 2.02 & 7100.3 & 0.31 & 0.44 & 9250.6 \\
 & ICA, DVA, V, Q & 0.26 & 0.29 & 16145.4 & 1.56 & 1.89 & 7730.9 & 0.16 & 0.19 & 10108.9 \\ \hline
\multirow{5}{*}{CNN-BiGRU}
 & ICA & 0.80 & 0.94 & 12064.4 & 0.45 & 0.71 & 5799.4 & 0.27 & 0.32 & 7509.2 \\
 & ICA, V & 0.47 & 0.53 & 13453.7 & 1.82 & 2.09 & 6521.3 & 0.19 & 0.22 & 8364.7 \\
 & ICA, V, Q & 0.51 & 0.65 & 1258.4 & 1.00 & 1.12 & 2559.1 & 0.19 & 0.27 & 3809.0 \\
 & ICA, DVA, V & 0.79 & 0.89 & 14866.8 & 0.86 & 1.25 & 7183.7 & 0.20 & 0.23 & 9264.1 \\
 & ICA, DVA, V, Q & 0.55 & 0.63 & 16142.5 & 0.83 & 1.05 & 3202.2 & 0.39 & 0.43 & 10156.4 \\ \hline
\end{tabular}
\end{adjustbox}
\label{SOH:cnn_rnn}
\end{table}
\begin{table}[!h]
\centering
\caption{Errors and FLOPS for the fourth category of the combination of CNN, FC and RNNs}
\begin{adjustbox}{width=1\columnwidth}
\begin{tabular}{|c|c|ccc|ccc|ccc|} \hline
\multicolumn{2}{|r}{Dataset} & \multicolumn{3}{c|}{Oxford} & \multicolumn{3}{c|}{NASA} & \multicolumn{3}{c|}{PoliMi} \\ \hline
Layer & Features & MAE (\%) & RMSE (\%) & FLOPs (k) & MAE (\%) & RMSE (\%) & FLOPS (k) & MAE (\%) & RMSE (\%) & FLOPS (k) \\ \hline
\multirow{5}{*}{CNN-D-LSTM}
 & ICA & 0.25 & 0.32 & 12174.85 & 1.04 & 1.33 & 3982.912 & 0.12 & 0.15 & 8605.504 \\
 & ICA, V & 0.19 & 0.23 & 2211.712 & 2.04 & 2.17 & 7425.152 & 0.24 & 0.28 & 9600.32 \\
 & ICA, V, Q & 0.27 & 0.30 & 10179.2 & 0.94 & 1.14 & 6478.528 & 0.33 & 0.47 & 7300.736 \\
 & ICA, DVA, V & 0.23 & 0.30 & 2980.608 & 0.77 & 1.27 & 8272.32 & 0.25 & 0.32 & 6586.688 \\
 & ICA, DVA, V, Q & 0.16 & 0.20 & 9379.712 & 1.38 & 1.57 & 2098.496 & 0.19 & 0.24 & 4149.952 \\\hline
\multirow{5}{*}{CNN-D-biLSTM}
 & ICA & 0.46 & 0.52 & 4258.048 & 1.58 & 1.98 & 8269.504 & 0.28 & 0.32 & 6424.448 \\
 & ICA, V & 0.16 & 0.19 & 15076.29 & 0.89 & 1.18 & 7374.656 & 0.47 & 0.51 & 9672.512 \\
 & ICA, V, Q & 0.17 & 0.21 & 4866.304 & 1.65 & 1.79 & 7927.744 & 0.47 & 0.56 & 5159.104 \\
 & ICA, DVA, V & 0.45 & 0.51 & 15347.07 & 1.34 & 1.46 & 7582.144 & 0.12 & 0.14 & 6424.192 \\
 & ICA, DVA, V, Q & 0.25 & 0.31 & 6664.64 & 0.78 & 1.04 & 2570.816 & 0.37 & 0.44 & 1184.96 \\\hline
\multirow{5}{*}{CNN-D-GRU}
 & ICA & 0.97 & 1.14 & 1181.824 & 1.12 & 1.26 & 6710.336 & 0.34 & 0.39 & 9778.24 \\
 & ICA, V & 0.71 & 0.86 & 2833.28 & 0.97 & 1.19 & 5116.544 & 0.38 & 0.43 & 1561.664 \\
 & ICA, V, Q & 0.36 & 0.43 & 9384.832 & 0.68 & 1.14 & 3202.624 & 0.18 & 0.23 & 414.784 \\
 & ICA, DVA, V & 0.35 & 0.45 & 5039.488 & 0.82 & 1.00 & 9565.504 & 0.24 & 0.28 & 5277.568 \\
 & ICA, DVA, V, Q & 0.65 & 0.73 & 3039.488 & 0.62 & 0.86 & 1374.144 & 0.18 & 0.22 & 6828.736 \\ \hline
\multirow{5}{*}{CNN-D-biGRU}
 & ICA & 1.19 & 1.31 & 12463.74 & 1.99 & 2.41 & 6549.184 & 0.53 & 0.61 & 3895.936 \\
 & ICA, V & 0.59 & 0.66 & 13823.1 & 1.27 & 1.56 & 5555.52 & 0.42 & 0.45 & 9438.784 \\
 & ICA, V, Q & 0.73 & 0.84 & 6489.024 & 2.83 & 3.01 & 7927.744 & 0.65 & 0.71 & 3857.152 \\
 & ICA, DVA, V & 0.28 & 0.31 & 16562.94 & 1.26 & 1.37 & 1425.984 & 0.48 & 0.53 & 7059.52 \\
 & ICA, DVA, V, Q & 0.59 & 0.73 & 17868.54 & 1.18 & 1.43 & 1304.384 & 0.47 & 0.60 & 10587.01 \\ \hline
\end{tabular}
\end{adjustbox}
\label{SOH:cnn_d_rnn}
\end{table}
\end{document}